\documentclass{article}
\usepackage[preprint]{tmlr}  % arXiv preprint (de-anonymized; removes 'Under review' banner)
\usepackage{amsmath,amssymb,booktabs,graphicx,xcolor}
\title{When the Tool Decides: LLM Agents Defer Blindly to Graph Neural Network Tools, and Stronger Backbones Defer More}
\author{\name Zhongyuan Wang \email zhongyuan@raptorx.ai \\
\addr raptorX.ai
\AND
\name Pratyusha Vemuri \email pratyusha@raptorx.ai \\
\addr raptorX.ai}
\date{}

\begin{document}
\maketitle

\begin{abstract}
A growing line of work equips large language model (LLM) agents with graph
neural networks (GNNs) and other structured predictors as callable tools, on
the implicit assumption that the agent exercises judgment over \emph{when} and
\emph{how much} to rely on such a tool. We test this assumption directly. We
expose a frozen GNN to a ReAct-style LLM agent as an explicit tool (returning a
predicted label, an anomaly score, and link probabilities) and measure, on
node classification over a text-attributed graph (ogbn-arxiv, with a
replication on WikiCS), whether the
agent uses the tool or merely obeys it. We find that the agent does not
exercise judgment: its predictions agree with the raw GNN's $97.6$--$99.2\%$ of
the time (5 seeds), i.e.\ the agent collapses into a \emph{GNN parrot} that
adopts the tool's output wholesale and bypasses its own reasoning. Sweeping
backbone capability (Qwen2.5 $0.5$B--$7$B) shows the deference is not a
weak-model artifact that capability removes; among models able to invoke the
tool at all, agreement \emph{increases} with capability ($0.60 \to 0.98$ from
$1.5$B to $7$B, 5 seeds). Crucially, the \emph{cost} of this deference shows no
shrinkage in any regime as capability grows, and grows significantly where
alternatives emerge: a per-node oracle that
selects the best of the available actions beats the parrot by $0.09$--$0.18$ at
$3$B and $0.12$--$0.22$ at $7$B, roughly doubling at high homophily
($0.12 \to 0.22$, positive in all $5$ paired seeds), because the parrot's
accuracy is pinned to the frozen GNN
while the agent's alternatives improve with capability: at $7$B a simple
neighbour-label lookup tool overtakes the GNN at high homophily ($0.81$ vs.\
$0.71$) yet the agent defers to the GNN regardless. A simple
selective-invocation gate recovers about half of that high-homophily gap
($0.71 \to 0.83$) but hurts elsewhere and yields no net global gain; moreover,
held-out estimates bound the best achievable gate over standard test-time
features to recovering at most about a third of the oracle headroom (the rest
appears not recoverable from those features), so reliable selective invocation
is an open problem that looks limited by available information, not merely
router design. Our results are a cautionary measurement for the
graph--LLM--agent and tool-augmented-agent communities: evaluations of
``agent~$+$~tool'' systems cannot assume the agent adds judgment on top of the
tool, and selective invocation must be designed in rather than expected to
emerge from scale.
\end{abstract}

\section{Introduction}
Tool-augmented LLM agents increasingly call learned models as black-box tools.
In the graph setting in particular, recent systems give an LLM agent access to
graph operations and learned graph models and report gains over the agent
alone. A premise underlying this design is that the agent behaves as a
\emph{discerning} caller: it should consult the tool when the tool is
trustworthy and fall back to other evidence (text, neighbourhood structure, its
own reasoning) when the tool is not. Whether agents actually behave this way has
not, to our knowledge, been measured head-to-head.

We ask a deliberately narrow, falsifiable question: \emph{when an LLM agent is
given a frozen GNN as an explicit tool, does it use the tool's output as one
piece of evidence, or does it simply obey it?} We operationalize ``obey'' with a
prediction-level \emph{agreement} between the agent's final answer and the raw
GNN prediction, and we operationalize the \emph{cost} of obeying with an
\emph{oracle gap}: how much a per-node oracle over the available actions would
have beaten the agent.

\paragraph{Who should care (audience).} This paper targets two TMLR
sub-audiences. (i) Researchers building graph--LLM agents or, more broadly,
tool-augmented agents that call \emph{learned} predictors: our measurement
shows that a common evaluation assumption (the agent contributes judgment over
the tool) can fail outright, which changes how such systems should be ablated
and reported. (ii) Researchers studying how agentic behavior scales with model
capability: we report a case where a desirable behavior (skeptical tool use)
does \emph{not} appear with scale and in fact the opposite trend holds. Neither
group needs the method to be novel to act on the finding; they need it to be
well-supported, which is our focus.

\paragraph{Contributions (as evidence, not novelty claims).}
\begin{itemize}
\item We provide evidence that an LLM agent given a frozen GNN tool collapses
  into a \emph{parrot}: prediction-level agreement with the raw GNN is
  $0.976$--$0.992$ across local-homophily regimes (Section~\ref{sec:parrot},
  Table~\ref{tab:main}), while agreement with its own tool-free reasoning is
  only $0.17$--$0.37$ ($7$B; $0.07$--$0.20$ at $3$B).
\item We show this deference is not removed by capability: across Qwen2.5
  $0.5$B--$7$B, once a model can use the tool at all ($\geq 1.5$B), agreement
  with the GNN \emph{increases} with capability, $0.60 \to 0.98$ (5 seeds)
  (Section~\ref{sec:capability}, Table~\ref{tab:capability}).
\item We show the \emph{cost} of deference shows no shrinkage in any regime as
  capability grows, and grows significantly where alternatives emerge: the
  per-node oracle gap is
  $0.09/0.18/0.12$ (low/mid/high) at $3$B and $0.12/0.18/0.22$ at $7$B,
  roughly doubling at high homophily (positive in all $5$ paired seeds, paired
  $t{=}9.1$), because
  the parrot is pinned to the frozen GNN while the alternatives strengthen: at
  $7$B the neighbour-label tool overtakes the GNN at high homophily
  ($0.81$ vs.\ $0.71$) yet the agent still defers
  (Section~\ref{sec:cost}, Table~\ref{tab:gap}).
\item We show a simple selective-invocation gate recovers about half the gap
  where its feature is informative (high homophily $0.71 \to 0.83$, oracle
  $0.93$; positive in all $5$ seeds) but hurts elsewhere, for no net global
  gain ($0.481 \to 0.475$), and a learned four-feature router does no better.
  An information-ceiling analysis sharpens this: two held-out estimators bound
  the \emph{best achievable} gate over standard test-time uncertainty features
  to recovering only one sixth to one third of the oracle headroom on arxiv
  (and $\approx 12$--$14\%$ on WikiCS), null-controlled, the rest appearing not
  recoverable from those features (Section~\ref{sec:gate}). Reliable selective
  invocation thus looks limited by available information, not merely router
  design, and remains an open problem.
\end{itemize}
We are explicit about scope (Section~\ref{sec:discussion}): results are on
ogbn-arxiv and replicated on WikiCS (Section~\ref{sec:wikics}) with the Qwen2.5
family; we do not claim the magnitudes transfer, we claim the failure mode
exists and is reproducible under controls.

\section{Related Work}
\paragraph{LLM agents that touch graphs.} Recent systems give agents
graph-native textual operations (neighbour lookup, $k$-hop retrieval) and
report improvements~\citep{agentgl}. These tools are textual; none, to our
knowledge, expose a \emph{frozen neural GNN} as an explicit tool whose output
the agent must decide to trust, which is exactly the object we measure. Our
navigation arm is a deliberately minimal neighbour-label lookup in the spirit
of (not reusing code from) \citet{agentgl}'s graph-native operations.
\paragraph{Do LLMs read graph structure?} \citet{whenstructure} show, in a
\emph{non-agentic} setting, that LLMs benefit little from structural encodings
beyond node text. Our question is distinct: not whether structure helps an LLM's
input, but whether an \emph{agent}, under a tool-calling budget, defers to a
structural tool. We include a pure-LLM arm to connect to their result.
\paragraph{Selective use of expensive components.} \citet{glance} learn
\emph{when to call an LLM} from a GNN's side; we study the mirror image (when an
agent should \emph{not} call/trust the GNN) and find agents fail to do it
unaided.
\paragraph{Capability vs.\ orchestration.} \citet{trankiela,scalingagents}
report that single strong models can match or beat multi-agent orchestration
under matched budgets, i.e.\ coordination gains shrink with capability. We
observe a related but distinct capability trend at the level of a single agent's
tool deference, and we resolve it specifically for the GNN-as-tool case.
\paragraph{Tool over-reliance and tool trust (concurrent).} A concurrent line
documents adjacent failure modes of tool-augmented LLMs:
\citet{toolmemory} study tool--memory conflicts in single-shot QA (the model
must arbitrate between a tool answer and parametric knowledge);
\citet{tooltax} report a cautionary ``tool-use tax'' (tool-augmented agents
underperforming plain CoT) with a lightweight gate that only partially recovers
it; \citet{reasoningtrap} find that \emph{strengthening} reasoning amplifies
tool hallucination, another capability-worsening trend; and \citet{toolposition}
argue agents should invoke tools only when epistemically necessary, a position
for which our measurement supplies graph-domain evidence. None of these expose
a frozen \emph{learned predictor} as a tool inside an agent loop and measure
prediction-level deference and its scaling with backbone capability, which is
our object. Despite the name, GNN-as-Judge \citep{gnnasjudge} uses GNN feedback
to filter pseudo-labels for LLM fine-tuning (training-time collaboration), not
a callable GNN tool at inference.

\section{Setup}
\label{sec:setup}
\paragraph{GNN-as-tool paradigm and arms.} A GCN is trained on the task and
\emph{frozen}; it is exposed to a ReAct LLM agent as a tool bound to the query
node, returning (E) a predicted label with confidence, (A) a reconstruction
anomaly score, and (L) link probabilities to neighbours. We compare four arms
under a matched per-query budget ($5{,}000$ prompt$+$generation tokens and $6$
tool calls per query):
\textbf{A1} agent~$+$~GNN-tool; \textbf{A2} agent~$+$~a minimal graph-navigation
tool in the spirit of \citet{agentgl}'s textual graph operations:
\texttt{neighbors()} returns up to $k{=}10$ neighbours with their training-set
labels where available (\emph{no neighbour text is exposed}) and
\texttt{degree()} the node degree;
\textbf{A3} the frozen GNN alone; \textbf{A4} the agent with no graph tool
(verbalized node text only).
\paragraph{Data and regimes.} We use ogbn-arxiv, a text-attributed citation
graph ($169$k nodes, $40$ classes) with the official title$+$abstract text as
node verbalization. We stratify test nodes by \emph{local homophily} (fraction
of same-label neighbours, ground truth) into low ($<0.3$), mid ($[0.3,0.7)$),
high ($\geq 0.7$); this is an analysis axis only and is never given to the agent.
\paragraph{Backbones and protocol.} Qwen2.5-Instruct at $0.5$B, $1.5$B, $3$B,
$7$B, served locally; agents use a text-protocol ReAct loop (regex-parsed
\texttt{ACTION}/\texttt{ANSWER} lines) with budget enforcement. Two protocol
facts matter for interpretation and are stated here rather than hidden: (i) the
scaffold \emph{instructs} the agent to consult tools before answering, so tool
\emph{invocation} is prompt-encouraged; our measurement is therefore about what
the agent does with the returned output (adopt vs.\ weigh), not about whether
it chooses to invoke, and the selective-invocation question of
Section~\ref{sec:gate} is posed \emph{outside} the agent for this reason;
(ii) the scaffold and instructions are in Chinese (Qwen2.5 is
Chinese--English bilingual; node text is English), which matters when probing
non-Qwen backbones (Section~\ref{sec:discussion}). Full prompts, decoding
parameters, budget and fallback rules are in Appendix~\ref{app:protocol}.
\paragraph{Metrics.} Accuracy per arm; \emph{agreement} $\mathrm{Pr}[\text{A1
pred}=\text{A3 pred}]$ as the deference (parrot) measure; \emph{oracle gap}
$\mathrm{acc}(\max\{A1,A2,A4\}) - \mathrm{acc}(A1)$ as the cost of deference (how
much a per-node best-action selector would beat the parrot). The gap is
non-negative by construction (the oracle's action set includes A1); its
informative content is its magnitude. Unless noted,
$7$B numbers are mean$\pm$SE over $5$ seeds (re-trained GNN $+$ resampled nodes),
$50$ nodes/bin. Seeds fix the GNN training and the node sample; LLM decoding is
sampled (temperature $0.7$) and not seed-fixed, so single-run numbers carry
decoding noise (Appendix~\ref{app:protocol}).
\paragraph{On the neighbour-label arm (not leakage).} A2 exposes neighbours'
\emph{training} labels (as a tool observation), never the query node's own test
label; this is the same supervision the GNN is trained on, so A2's strength at
high homophily reflects legitimate use of training signal (label voting over an
informative neighbourhood), not test leakage. Conversely, its weakness at low
homophily is intrinsic to label voting in disassortative neighbourhoods; an A2
variant with neighbour-\emph{text} access might behave differently and is left
to future work.

\section{The agent parrots the GNN}
\label{sec:parrot}
Table~\ref{tab:main} ($7$B) shows A1 (agent$+$GNN-tool) is operationally
indistinguishable from A3 (raw GNN): prediction agreement is $0.976$--$0.992$
across regimes. The agent calls the tool roughly once and adopts its label.
Because the scaffold encourages tool use (Section~\ref{sec:setup}), the
invocation itself partly reflects compliance; the parrot finding is about what
happens \emph{after} the call. Two observations sharpen it. First, the same
agent's tool-using answers coincide with its own tool-free reasoning (A4) only
$17$--$37\%$ of the time ($7$--$20\%$ at $3$B): the tool, once available,
overrides the agent's own reasoning almost entirely. Second, the toolbox
exposes \emph{three} signals as separate calls (predicted label with
confidence; an anomaly score that flags nodes where the GNN is likely wrong;
link probabilities), with budget for $6$ calls, yet in $83\%$ of $7$B queries
the agent makes exactly one call: it reads the label and never probes the very
signal designed to flag when the label should not be trusted.
We call this collapse a \emph{GNN parrot}.

\begin{table}[t]
\centering
\caption{$7$B (Qwen2.5-7B), ogbn-arxiv, by local homophily. Mean$\pm$SE over
$5$ seeds, $50$ nodes/bin. A1$\approx$A3 (agreement column) is the parrot
effect; the oracle gap is the cost of deferring. A2 (neighbour-label) overtakes
the GNN at high homophily (and is comparable at mid).}
\label{tab:main}
\begin{tabular}{lccccccc}
\toprule
homophily & A1 parrot & A2 nbr-label & A3 GNN & A4 no-tool & oracle & oracle gap & agree(A1$=$A3) \\
\midrule
low ($<0.3$)   & $0.29\pm0.04$ & $0.13\pm0.02$ & $0.30\pm0.04$ & $0.16\pm0.03$ & $0.41\pm0.04$ & $0.120\pm0.006$ & $0.976\pm0.007$ \\
mid            & $0.44\pm0.03$ & $0.40\pm0.04$ & $0.44\pm0.03$ & $0.20\pm0.03$ & $0.62\pm0.02$ & $0.184\pm0.037$ & $0.976\pm0.012$ \\
high ($\geq0.7$) & $0.71\pm0.05$ & $0.81\pm0.03$ & $0.71\pm0.05$ & $0.42\pm0.02$ & $0.93\pm0.02$ & $0.220\pm0.030$ & $0.992\pm0.005$ \\
\bottomrule
\end{tabular}
\end{table}

\begin{figure}[t]
\centering
\includegraphics[width=\textwidth]{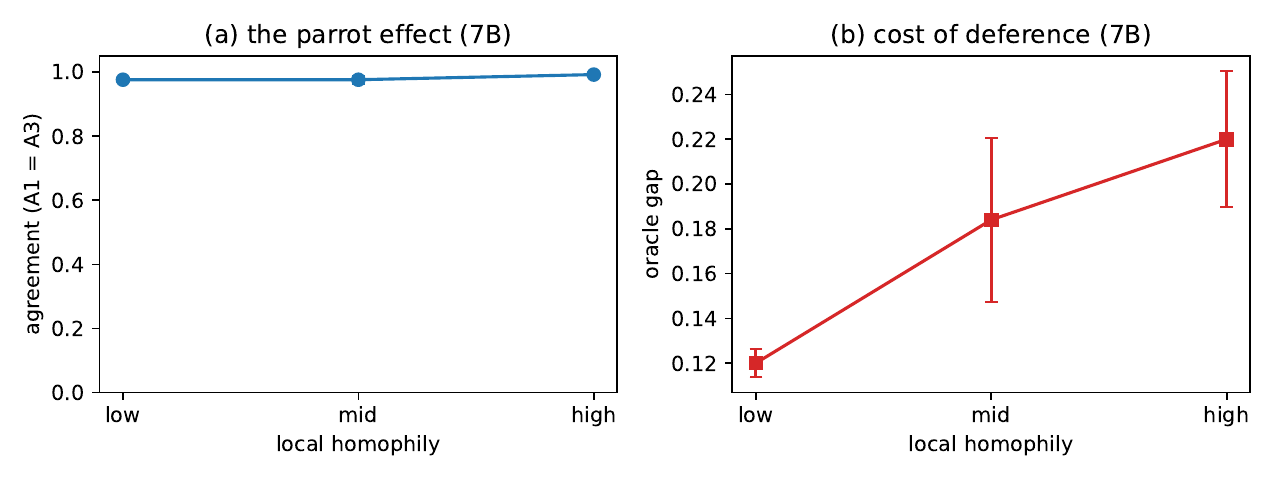}
\caption{$7$B, mean$\pm$SE over $5$ seeds. (a) Agreement (A1$=$A3) stays near $1$
across local-homophily regimes (the parrot effect). (b) The oracle gap (cost of
deference) is positive throughout.}
\label{fig:parrot}
\end{figure}

\section{Capability deepens deference}
\label{sec:capability}
Table~\ref{tab:capability} sweeps backbone capability. At $0.5$B the model
cannot reliably use the tool at all (it issues almost no valid tool calls; low
agreement here is incapacity, not skepticism). From $1.5$B upward, where the
agent does invoke the tool, agreement with the GNN \emph{rises} with capability
and saturates near $1$: averaging over bins, $0.60 \to 0.97 \to 0.98$ for
$1.5$/$3$/$7$B (5 seeds). Capability does not buy skepticism;
it buys more complete deference.

\begin{table}[t]
\centering
\caption{Deference vs.\ backbone capability. Agreement (A1$=$A3) per
local-homophily bin: $1.5$/$3$/$7$B are mean$\pm$SE over $5$ seeds
($50$ nodes/bin); the $0.5$B row and the calls column come from the seed-$0$
capability run ($30$ nodes/bin). ``calls'' is the mean tool-call count at low
homophily, distinguishing incapacity (calls$\approx$0) from deference. When a
model emits no parseable answer the harness falls back to class $0$
(Appendix~\ref{app:protocol}), so the $0.5$B row reflects incapacity rather
than skepticism.}
\label{tab:capability}
\begin{tabular}{lcccc}
\toprule
backbone & agree low & agree mid & agree high & tool calls (low) \\
\midrule
Qwen2.5-0.5B & $0.10$ & $0.00$ & $0.10$ & $0.30$ \;(cannot use tool) \\
Qwen2.5-1.5B & $0.62{\pm}.02$ & $0.60{\pm}.03$ & $0.58{\pm}.05$ & $1.23$ \\
Qwen2.5-3B   & $0.98{\pm}.00$ & $0.96{\pm}.01$ & $0.98{\pm}.01$ & $1.00$ \\
Qwen2.5-7B   & $0.98{\pm}.01$ & $0.98{\pm}.01$ & $0.99{\pm}.01$ & $1.07$ \\
\bottomrule
\end{tabular}
\end{table}

\begin{figure}[t]
\centering
\includegraphics[width=0.5\textwidth]{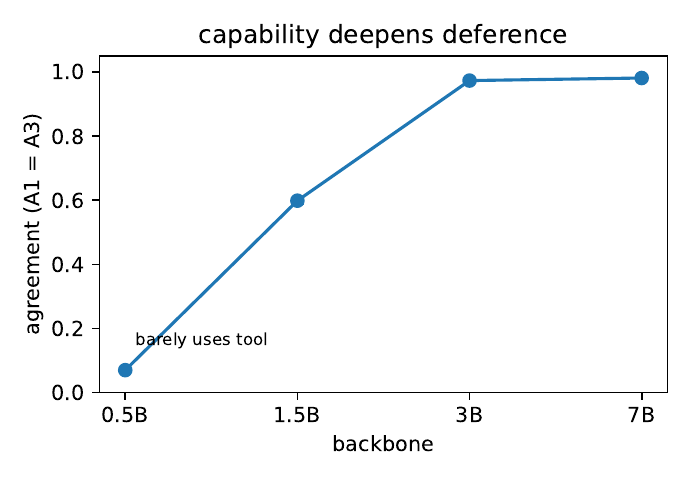}
\caption{Agreement with the GNN rises with backbone capability among tool-using
models ($1.5$B+); the $0.5$B model barely calls the tool. Mean$\pm$SE.}
\label{fig:capability}
\end{figure}

\section{The cost of deference does not shrink with capability}
\label{sec:cost}
Deference is harmless if the tool is always best. It is not, and capability
makes it worse where it matters most. Table~\ref{tab:gap} reports the per-node
oracle gap across backbones, on the same sampled nodes per seed. Two
regularities emerge. (i) From $3$B to $7$B we observe no regime where the gap
shrinks, and significant growth where alternatives emerge: it
roughly doubles at high homophily ($0.12 \to 0.22$; the paired per-seed
difference is positive in all $5$ seeds, paired $t{=}9.1$, which survives a
Bonferroni correction across the three bins), is directionally larger at low
($0.09 \to 0.12$; $4$ of $5$ paired seeds, $t{=}0.8$, not significant), and
is unchanged at mid ($0.18$, $t{=}0.0$). (ii) The mechanism is visible in the arms: the
parrot's accuracy is pinned to the frozen GNN (bin-mean A1 within $\pm 0.02$ of
A3 for $3$B/$7$B), while the alternatives strengthen with
capability: the tool-free arm A4 rises from at most $0.10$ ($1.5$B) to
$0.16$--$0.42$ ($7$B), and at $7$B (and only $7$B) the neighbour-label arm
\emph{overtakes} the GNN at high homophily ($0.81$ vs.\ $0.71$,
Table~\ref{tab:main}), because a capable agent can aggregate neighbours'
training labels when the neighbourhood is informative. At mid homophily the two
arms are comparable ($0.40$ vs.\ $0.44$) and the $0.18$ gap reflects per-node
complementarity rather than a single dominant alternative. The $1.5$B row shows
the complementary failure mode: an agent too weak to even match its tool
(agreement $0.60$; A1 trails the GNN by $0.12$--$0.28$) pays for its
\emph{deviations}, so its gap is not a pure cost of deference, which is why we
state the capability claim over the full-parrot regime ($3$B/$7$B). There, the
stronger the agent, the more it leaves on the table by obeying the GNN,
precisely because it had better alternatives (the navigation tool, or its own
reasoning) available and unused.

\begin{table}[t]
\centering
\caption{Per-node oracle gap (cost of deference) vs.\ backbone capability,
ogbn-arxiv, mean$\pm$SE over $5$ seeds ($50$ nodes/bin; node samples are shared
across backbones within a seed, so the $3$B-vs-$7$B comparison is paired). The
$3$B and $7$B agents are full parrots (agreement $\geq 0.96$); the $1.5$B agent
defers only partially (agreement $0.60$) and its deviations from the GNN cost
accuracy, so its gap mixes deference with incompetent deviation.}
\label{tab:gap}
\begin{tabular}{lccc}
\toprule
backbone & gap low & gap mid & gap high \\
\midrule
Qwen2.5-1.5B & $0.048\pm.010$ & $0.168\pm.033$ & $0.228\pm.014$ \\
Qwen2.5-3B   & $0.092\pm.032$ & $0.184\pm.024$ & $0.120\pm.033$ \\
Qwen2.5-7B   & $0.120\pm.006$ & $0.184\pm.037$ & $0.220\pm.030$ \\
\bottomrule
\end{tabular}
\end{table}

\section{A selective-invocation gate, and its limits}
\label{sec:gate}
If the failure is \emph{indiscriminate} deference, the remedy is to gate it. We
test a simple post-hoc gate that routes each node to A2 (neighbour-label) when
the purity of its training-label neighbourhood exceeds a threshold
($\tau{=}0.4$, chosen on seed $0$) and to A1 (GNN) otherwise, using only
test-time--available information, evaluated over the same $5$ seeds as
Table~\ref{tab:main}. Where its feature is informative the gate recovers about
half the remaining gap: at high homophily it lifts $0.71 \to 0.83$ (oracle
$0.93$; paired $t{=}4.0$), positive in all $5$ seeds, including the $4$ seeds
unseen by the $\tau$ choice. But it \emph{hurts} where the purity proxy
is unreliable (low $0.29 \to 0.18$, mid $0.44 \to 0.41$), for no net global
gain ($0.481 \to 0.475$, in fact slightly negative). (A seed-$0$-only
evaluation had suggested a global
$+0.07$ gain; it does not survive $5$ seeds, which we report as a caution
against single-seed gate evaluations.) We further train a learned router over
four features (purity, GNN-confidence, degree, neighbour-disagreement),
routing each node to \{GNN, A2, A4\}. Evaluated leave-one-seed-out over the
$750$ stratified nodes of Table~\ref{tab:main} (train on four seeds, test on
the fifth), the learned router ($0.496\pm0.026$) does \emph{not} beat a purity
gate whose threshold is validation-selected on the training seeds
($0.499\pm0.016$), and neither meaningfully improves on the parrot
($0.481\pm0.026$) against a per-node oracle of $0.656\pm0.024$. An earlier
iid-sampled single-split run ($n{=}300$, $150$ held-out) reached the same
conclusion ($0.527$ vs.\ $0.533$, oracle $0.647$).

Is this a failure of our particular routers, or of the available information?
To bound the \emph{best achievable} gate over these four features we estimate
it held-out two ways: a leave-one-seed-out $k$-nearest-neighbour best-arm
policy ($k{=}31$ in standardized feature space) reaches $0.537\pm0.025$, and a
coarser median-binned cell policy reaches $0.508\pm0.021$; both clear their
feature-shuffled nulls ($0.469$ and $0.472$; real $>$ null in all $5$ seeds)
but sit far below the per-node oracle ($0.656$). Together they bound the share
of the $0.175$ oracle headroom recoverable from these features to roughly
\emph{one sixth to one third} ($0.027$--$0.056$ above the parrot): the majority
is not recovered, and the residual gap to the oracle ($\approx 0.12$--$0.15$)
is not closed by any gate we could build over these features. This is an
empirical ceiling, not a proof that no feature could help; it says the standard
uncertainty proxies (purity, GNN-confidence, degree, neighbour-disagreement)
are insufficient. The binding constraint thus looks like the information
available at test time, not the router class; our simpler gates capture even
less. The same analysis on WikiCS recovers an even smaller share
($\approx 12$--$14\%$, Section~\ref{sec:wikics}), so this is not arxiv-specific.
Reliably knowing when to distrust the GNN is thus an \emph{open problem}
that appears, in part, \emph{informational}: closing it likely needs signals
beyond standard uncertainty proxies. We present selective invocation as a
necessary direction whose realization remains open.

\section{Replication on a second graph (WikiCS)}
\label{sec:wikics}
To check the findings are not specific to ogbn-arxiv, we replicate the $7$B
measurement on WikiCS (Wikipedia computer-science articles; $11.7$k nodes,
$10$ classes, edge homophily $0.66$; a different domain at comparable
homophily), $3$ seeds. The parrot effect holds: agreement (A1$=$A3) is
$0.96$--$1.00$ (Table~\ref{tab:wikics}), so the agent again adopts the GNN
wholesale. The oracle gap is again positive in every bin of every seed ($9/9$
seed$\times$bin pairs; $0.03$--$0.23$), so deference again costs accuracy. The \emph{regime} where the cost peaks differs
(on WikiCS the gap is largest at low homophily, $0.23$, where the
neighbour-label arm overtakes the GNN, $0.38$ vs.\ $0.25$; on arxiv it peaked
at high homophily):
\emph{which} alternative beats the GNN is dataset-dependent, but the qualitative
findings (indiscriminate deference and a positive oracle gap) reproduce.
The information-ceiling analysis of Section~\ref{sec:gate} also reproduces: on
WikiCS the best achievable held-out gate over the same four features recovers
only $0.014$--$0.017$ of the $0.119$ oracle headroom (kNN and cell estimators;
$\approx 12$--$14\%$, again clearing its feature-shuffled null in all $3$ seeds
but by a small margin), so the bulk of the deference cost is again not
recoverable from standard uncertainty features---if anything more locked than on
arxiv.

\begin{table}[t]
\centering
\caption{WikiCS, $7$B, mean$\pm$SE over $3$ seeds, $40$ nodes/bin. Parrot
(agreement $0.96$--$1.00$) and a positive oracle gap reproduce the arxiv
findings in a different domain.}
\label{tab:wikics}
\begin{tabular}{lccccc}
\toprule
homophily & parrot & nbr-label & oracle & oracle gap & agreement \\
\midrule
low  & $0.25\pm0.03$ & $0.38\pm0.03$ & $0.48\pm0.03$ & $0.23\pm0.01$ & $0.98\pm0.02$ \\
mid  & $0.75\pm0.03$ & $0.68\pm0.04$ & $0.85\pm0.03$ & $0.10\pm0.01$ & $0.96\pm0.02$ \\
high & $0.96\pm0.02$ & $0.85\pm0.03$ & $0.99\pm0.01$ & $0.03\pm0.01$ & $1.00\pm0.00$ \\
\bottomrule
\end{tabular}
\end{table}

\section{Discussion: scope and limitations}
\label{sec:discussion}
Our claims are scoped to ogbn-arxiv and WikiCS node classification with the
Qwen2.5 family and a GCN tool; we do not claim the magnitudes transfer. We claim
the failure mode (indiscriminate deference; its worsening with capability; the
resulting oracle gap) exists and is reproducible under controls, and we
replicate the parrot effect and a positive oracle gap on a second graph in a
different domain (Section~\ref{sec:wikics}). The capability sweep is $5$-seed
for $1.5$/$3$/$7$B ($0.5$B is seed 0, as it barely uses the tool); the
\emph{agreement} trend saturates by $3$B ($0.97 \to 0.98$ from $3$B to $7$B is
within noise), so the load-bearing agreement contrast is $1.5$B vs.\ $3$B$+$,
while the \emph{cost} contrast ($3$B vs.\ $7$B, Table~\ref{tab:gap}) does not
saturate; backbones beyond $7$B under the same protocol are left to future
work (single-GPU constraint). The capability
trend is observational: we do not isolate a mechanism (e.g., the GNN tool's
output dominating the agent's context), which we leave open. Neither of our
gates (a single hand-designed feature; a small learned router) closes the gap
globally; richer routers remain open.

\paragraph{Cross-family: a boundary condition.} We probed two non-Qwen families.
Under our original Chinese scaffold, Mistral-7B-Instruct rarely invoked the tool
($0.39$ calls, $n{=}40$) and showed no parrot effect (agreement $0.20$),
confounding ``does not parrot'' with ``does not use the tool''. Re-running under
a matched \emph{English} scaffold removes this confound: on a shared frozen GNN
and node sample ($3$ seeds, $60$ nodes), Mistral-7B and OLMo-2-7B-Instruct both
invoke the tool readily ($1.27$ and $1.39$ calls). The control is essential:
Qwen-7B under the same English scaffold still parrots (agreement
$0.978\pm0.015$, matching its $0.98$ under Chinese), so the scaffold language is
not what produces parroting. Yet the two other families defer only
\emph{partially}: agreement with the GNN is $0.53\pm0.01$ (Mistral) and
$0.60\pm0.03$ (OLMo), far below Qwen's $0.98$, and their deviations cost
accuracy (A1 $0.32$/$0.36$ vs.\ the shared GNN's $0.49$). So near-total
parroting is, on this evidence, partly \emph{Qwen-specific}: every tool-using
agent we tested agrees with the GNN on a majority of nodes, but the wholesale
collapse (agreement $\geq 0.97$) is strongest in Qwen. We state the strong-parrot
claim for the Qwen family and report this as a boundary condition---the effect's
direction generalizes across families, its extreme magnitude does not---and note
that reliable tool-use is a \emph{precondition} for any parrot effect (the
$0.5$B Qwen and Chinese-scaffold Mistral, which barely call the tool, do not
parrot).
Heterophilous text-attributed graphs (where the GNN is weak by construction)
would further stress the cost axis and are also left to future work.
None of these qualifications affect the core, error-barred results in
Tables~\ref{tab:main}--\ref{tab:wikics}.

\section{Conclusion}
Giving an LLM agent a GNN as a tool does not yield a discerning user of that
tool; it yields a parrot that adopts the tool's output wholesale, more
completely as the backbone grows stronger, at a cost that does not shrink with
capability, because stronger agents forgo better alternatives that their own
capability created. Evaluations of agent$+$learned-tool
systems should not assume the agent contributes judgment, and selective
invocation should be engineered rather than expected to emerge from scale.

\paragraph{Reproducibility.} Frozen-GNN training, the four arms, budget
enforcement, and all metrics are released; results JSON and seeds are provided.

\section*{Broader Impact}
This is a measurement study of when LLM agents over-trust a learned tool; it
introduces no new deployable system. Its intended impact is cautionary: agent
$+$ tool pipelines can silently inherit a tool's errors when the agent defers
indiscriminately, which is most consequential in high-stakes settings (fraud
detection, content moderation, scientific screening) where the tool may be
unreliable on tail inputs. We use only public benchmarks (ogbn-arxiv, WikiCS)
and open-weight models; no human subjects or private data are involved.

\bibliographystyle{plainnat}
\bibliography{references}

\begin{thebibliography}{10}
\providecommand{\natexlab}[1]{#1}
\providecommand{\url}[1]{\texttt{#1}}
\expandafter\ifx\csname urlstyle\endcsname\relax
  \providecommand{\doi}[1]{doi: #1}\else
  \providecommand{\doi}{doi: \begingroup \urlstyle{rm}\Url}\fi

\bibitem[Cheng et~al.(2026)Cheng, Pan, and Amiri]{toolmemory}
Jiali Cheng, Rui Pan, and Hadi Amiri.
\newblock Investigating tool-memory conflicts in tool-augmented llms, 2026.
\newblock URL \url{https://arxiv.org/abs/2601.09760}.

\bibitem[Kim et~al.(2026)Kim, Gu, Park, Park, Schmidgall, Heydari, Yan, Zhang,
  Zhuang, Liu, Malhotra, Liang, Park, Yang, Xu, Du, Patel, Althoff, McDuff, and
  Liu]{scalingagents}
Yubin Kim, Ken Gu, Chanwoo Park, Chunjong Park, Samuel Schmidgall, A.~Ali
  Heydari, Yao Yan, Zhihan Zhang, Yuchen Zhuang, Yun Liu, Mark Malhotra,
  Paul~Pu Liang, Hae~Won Park, Yuzhe Yang, Xuhai Xu, Yilun Du, Shwetak Patel,
  Tim Althoff, Daniel McDuff, and Xin Liu.
\newblock Towards a science of scaling agent systems, 2026.
\newblock URL \url{https://arxiv.org/abs/2512.08296}.

\bibitem[Loveland et~al.(2025)Loveland, Yang, and Koutra]{glance}
Donald Loveland, Yao-An Yang, and Danai Koutra.
\newblock Glance for context: Learning when to leverage llms for node-aware
  gnn-llm fusion, 2025.
\newblock URL \url{https://arxiv.org/abs/2510.10849}.

\bibitem[Sun et~al.(2026)Sun, Li, Fan, Liu, and Tan]{agentgl}
Yuanfu Sun, Kang Li, Dongzhe Fan, Jiajin Liu, and Qiaoyu Tan.
\newblock Agentgl: Towards agentic graph learning with llms via reinforcement
  learning, 2026.
\newblock URL \url{https://arxiv.org/abs/2604.05846}.

\bibitem[Tran and Kiela(2026)]{trankiela}
Dat Tran and Douwe Kiela.
\newblock Single-agent llms outperform multi-agent systems on multi-hop
  reasoning under equal thinking token budgets, 2026.
\newblock URL \url{https://arxiv.org/abs/2604.02460}.

\bibitem[Wang et~al.(2026)Wang, Qian, Li, Qiu, Xue, Wang, Ji, Storkey, and
  Wong]{toolposition}
Hongru Wang, Cheng Qian, Manling Li, Jiahao Qiu, Boyang Xue, Mengdi Wang, Heng
  Ji, Amos Storkey, and Kam-Fai Wong.
\newblock Position: Agent should invoke external tools only when epistemically
  necessary, 2026.
\newblock URL \url{https://arxiv.org/abs/2506.00886}.

\bibitem[Xu et~al.(2026)Xu, You, and Ma]{whenstructure}
Haotian Xu, Yuning You, and Tengfei Ma.
\newblock When structure doesn't help: Llms do not read text-attributed graphs
  as effectively as we expected, 2026.
\newblock URL \url{https://arxiv.org/abs/2511.16767}.

\bibitem[Xu and Ding(2026)]{gnnasjudge}
Ruiyao Xu and Kaize Ding.
\newblock Gnn-as-judge: Unleashing the power of llms for graph learning with
  gnn feedback, 2026.
\newblock URL \url{https://arxiv.org/abs/2604.08553}.

\bibitem[Yin et~al.(2026)Yin, Sha, Cui, Meng, and Li]{reasoningtrap}
Chenlong Yin, Zeyang Sha, Shiwen Cui, Changhua Meng, and Zechao Li.
\newblock The reasoning trap: How enhancing llm reasoning amplifies tool
  hallucination, 2026.
\newblock URL \url{https://arxiv.org/abs/2510.22977}.

\bibitem[Zhang et~al.(2026)Zhang, Xiong, Zhong, Jiang, Yuan, Li, and
  Lin]{tooltax}
Kaituo Zhang, Zhen Xiong, Mingyu Zhong, Zhimeng Jiang, Zhouyuan Yuan, Zhecheng
  Li, and Ying Lin.
\newblock Are tools all we need? unveiling the tool-use tax in llm agents,
  2026.
\newblock URL \url{https://arxiv.org/abs/2605.00136}.

\end{thebibliography}

\appendix
\section{Protocol details}
\label{app:protocol}
All scaffold prompts are in Chinese (we give faithful English translations
here; the verbatim originals ship in the supplementary code,
\texttt{src/agent.py} and \texttt{src/arms.py}). Node text (title$+$abstract,
truncated to $500$ characters) is English and is embedded verbatim in the
Chinese scaffold.

\paragraph{Agent system prompt (A1/A2), translated.} ``You are a
classification agent; call tools to gather evidence first, then answer.
Available tools: \texttt{<tool spec>}. At each step output exactly one line:
`ACTION: tool(args)' to call a tool, or `ANSWER: <class id 0..C$-$1>' to
answer.'' Note the prompt \emph{instructs} tool consultation
(Section~\ref{sec:setup}): invocation is compliance; adoption is the measured
behavior.

\paragraph{Task template (all arms), translated.} ``Paper:
`\texttt{<title+abstract, $\leq$500 chars>}'. Which class does this node belong
to? Candidates: \texttt{<id: label-name list>}.''
The A4 system prompt is ``You are a node classifier. Output exactly one line
`ANSWER: <class id>'\,''.

\paragraph{Tools.} A1 exposes three separate calls bound to the query node:
\texttt{gnn\_predict()} $\to$ ``class=$c$ conf=$p$'';
\texttt{gnn\_anomaly()} $\to$ a reconstruction anomaly score (described to the
agent as ``higher means the GNN is more likely wrong on this node'');
\texttt{gnn\_link(neighbour-id)} $\to$ a link probability. A2 exposes
\texttt{neighbors()} $\to$ up to $k{=}10$ neighbour ids, each with its training
label if the neighbour is in the training set (``?'' otherwise; no text), and
\texttt{degree()}.

\paragraph{Loop, budget, fallback.} ReAct loop of at most $4$ steps; each tool
observation is fed back as a user turn. Per-query budget: $5{,}000$
prompt$+$generation tokens (counted with the backbone's own tokenizer) and $6$
tool calls; on exhaustion, or after the last step, a forced-finalization turn
asks for `ANSWER:' based on the evidence so far. If the final output still
contains no in-range class id (the parser accepts an `ANSWER:' line or,
failing that, the last in-range integer in the text), the harness falls back
to class $0$ (this
affects only models that fail to follow the format, in practice $0.5$B and
Mistral). A1/A2/A4 share the same budget accounting; A3 charges one call.

\paragraph{Decoding and serving.} temperature $0.7$, top-$p$ $0.9$, max $256$
new tokens per turn. Seeds fix GNN training and node sampling, \emph{not} LLM
decoding, so repeated runs of the same seed differ by decoding noise; all core
claims are therefore stated over $5$ seeds (Mistral: $3$ independent runs).
Qwen $1.5$--$7$B multiseed runs are served by vLLM (behavior verified
equivalent to HuggingFace \texttt{transformers} on the seed-$0$ phase-1 grid);
$0.5$B and Mistral run on \texttt{transformers}.

\paragraph{GNN tool.} $2$-layer GCN (hidden $128$) with a reconstruction head,
trained $200$ epochs on the official training split of each dataset with
best-validation checkpoint selection, then frozen; per-seed retraining.
\end{document}